\documentclass{esannV2}
\usepackage{graphicx}
\usepackage[latin1]{inputenc}
\usepackage{amssymb,amsmath,array}
\usepackage{subfig}
 \usepackage{url} 
 
 \usepackage{listings}
\usepackage{float}
\usepackage{placeins}
\usepackage{tabularx}
\usepackage{textpos}

\newcommand{\R}{\mathbb{R}}

\newcommand{\y}{\textbf{y}}
\renewcommand{\u}{\textbf{u}}
\newcommand{\x}{\textbf{x}}
\newcommand{\W}{\textbf{W}}
\newcommand{\Win}{\W_{in}}

\newcommand{\Wout}{\W_{out}}

\begin{document}
\title{Comparison between DeepESNs and gated RNNs on multivariate time-series prediction}

\author{Claudio Gallicchio and Alessio Micheli and Luca Pedrelli
\vspace{.3cm}\\
Department of Computer Science, University of Pisa\\
Largo Bruno Pontecorvo 3 - 56127 Pisa, Italy\\
}

\maketitle
\begin{textblock*}{150mm}(-2cm,-6cm)
\emph{This is a preprint version of the following paper, published in the proceedings of ESANN 2019:} \\ Claudio Gallicchio, Alessio Micheli and Luca Pedrelli (2019) Comparison between DeepESNs and gated RNNs on multivariate time-series prediction. In: ESANN 2019 proceedings, European Symposium on Artificial Neural Networks, Computational Intelligence and Machine Learning. Bruges (Belgium), 24-26 April 2019, i6doc.com publ., ISBN 978-287-587-065-0.
\end{textblock*}
\begin{abstract}
We propose an experimental comparison between  Deep Echo State Networks (DeepESNs) and gated Recurrent Neural Networks (RNNs) on multivariate time-series prediction tasks. In particular, we compare reservoir and fully-trained RNNs able to represent signals featured by multiple time-scales dynamics. The analysis is performed in terms of efficiency and prediction accuracy on 4 polyphonic music tasks. Our results show that DeepESN is able to outperform ESN in terms of prediction accuracy and efficiency. Whereas, between fully-trained approaches, Gated Recurrent Units (GRU) outperforms Long Short-Term Memory (LSTM) and simple RNN models in most cases. Overall, DeepESN turned out to be extremely more efficient than others RNN approaches and the best solution in terms of prediction accuracy on 3 out of 4 tasks.

\end{abstract}

\section{Introduction}

Recurrent Neural Networks (RNNs) are a class of neural networks suitable for time-series processing. In particular, gated RNNs \cite{hochreiter1997long,cho2014learning}, such as Long Short-Term Memory (LSTM) and Gated Recurrent Units (GRU), are fully-trained recurrent models that implement adaptive gates able to address signals characterized by multiple time-scales dynamics.
Recently, within the Reservoir Computing (RC) \cite{lukovsevivcius2009reservoir} framework, the Deep Echo State Network (DeepESN) model has been proposed as extremely efficient way to design and training of deep neural networks for temporal data, with the intrinsic ability to represent hierarchical and distributed temporal features \cite{gallicchio2017deep, gallicchio2017hierarchical, DesignDeepESN}.

In this paper, we 
investigate different approaches to RNN modeling (i.e., untrained stacked layers and fully-trained gated architectures),
through an experimental comparison between RC and fully-trained RNNs on challenging real-world prediction tasks characterized by multivariate time-series. In particular, we perform a comparison between DeepESN, LSTM and GRU models on 4 polyphonic music tasks \cite{boulanger2012modeling}. Since these datasets are characterized by sequences with high-dimensionality and complex temporal sequences, these challenging tasks are particularly suitable for RNNs evaluation \cite{bengio2013advances}. Moreover, we consider ESN and simple RNN (Simple Recurrent Network - SRN) as baseline approaches for DeepESN and gated RNNs, respectively. The models are evaluated in terms of predictive accuracy and computation efficiency.

In a context in which the model design is difficult, especially for fully-trained RNNs, this paper would provide a first glimpse in the experimental comparison between different state-of-the-art recurrent models on multivariate time-series prediction tasks which still lacks in literature.

\section{Deep Echo State Networks}
\label{sec.DeepESN}
\begin{figure}[h]
\center
    \includegraphics[width=1\textwidth]{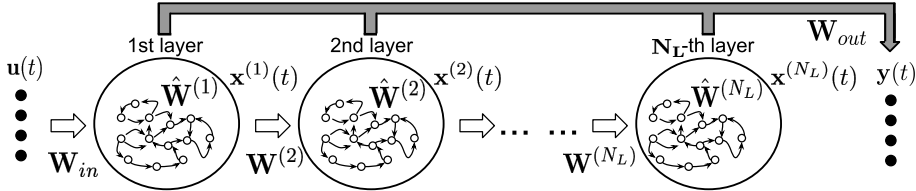}
    \caption{Hierarchical architecture of DeepESN.}
\label{fig:DeepESN}
\end{figure}

DeepESNs 
\cite{gallicchio2017deep}
extend Echo State Network (ESN) \cite{Jaeger2004} models
to the deep learning paradigm. Fig.~\ref{fig:DeepESN} shows an example of a DeepESN architecture composed by a hierarchy of $N_L$ reservoirs, coupled by a readout output layer.

In the following equations, $\u(t) \in \R^{N_U}$ 
and $\x^{(l)}(t) \in \R^{N_R}$ represent the external input and state of the l-th reservoir layer at step $t$, respectively.
Omitting bias terms for the ease of notation, and using leaking-rate reservoir units, the state transition of the first recurrent layer is described as follows:
\begin{equation}
\label{eq.layer1}
\x^{(1)}(t) = (1-a^{(1)}) \x^{(1)}(t-1) + a^{(1)} \mathbf{f}(\Win \u(t) + \hat{\W}^{(1)} \x^{(1)}(t-1)),
\end{equation}
while for each layer $l>1$ the state computation is performed as follows:
\begin{equation}
\label{eq.layeri}
\x^{(l)}(t) = (1-a^{(l)}) \x^{(l)}(t-1) + a^{(l)} \mathbf{f}(\W^{(l)} \x^{(l-1)}(t) + \hat{\W}^{(l)} \x^{(l)}(t-1)).
\end{equation}
In eq. \ref{eq.layer1} and \ref{eq.layeri}, $\Win \in \R^{N_R \times N_U}$ represents the matrix of input weights, $\hat{\W}^{(l)} \in \R^{N_R \times N_R}$ is the matrix of the recurrent weights of layer $l$, $\W^{(l)} \in \R^{N_R \times N_R}$ is the matrix that collects the inter-layer weights
from layer $l-1$ to layer $l$, $a^{(l)}$ is the leaky parameter at layer $l$ and $\mathbf{f}$ is the activation function of recurrent units implemented by a hyperbolic tangent ($\mathbf{f}\equiv\mathbf{tanh}$). Finally, the (global) state of the DeepESN is given by the concatenation of all the states encoded in the recurrent layers of the architecture $\x(t) = (\x^{(1)}(t), 
\ldots, \x^{(N_L)}(t)) \in \R^{N_L N_R}$. 

The weights in matrices $\W_{in}$ and $\{\W^{(l)}\}_{l = 2}^{N_L}$ are randomly initialized from a uniform distribution and re-scaled such that $\lVert \W_{in} \rVert{_2} = \sigma$ and $\lVert \W^{(l)} \rVert{_2} = \sigma$ respectively, where $\sigma$ is 
an input scaling parameter.
Recurrent layers are initialized in order to satisfy the necessary condition for the Echo State Property of DeepESNs \cite{gallicchio2017echo}. Accordingly, values in $\{\hat{\W}^{(l)}\}_{l=1}^{N_L}$ are randomly initialized from uniform distribution and re-scaled such that  $\max_{1 \leq l \leq N_L} \rho\left((1-a^{(l)}) \mathbf{I} +  a^{(l)}\hat{\W}^{(l)}\right) < 1$, 
where $\rho$ is the spectral radius of its matrix argument, i.e. the maximum among its eigenvalues in modulus.
The standard ESN case is obtained considering DeepESN with 1 single layer, i.e. when $N_L = 1$.

The output of the network at time-step $t$ 
is computed by the readout as a linear combination 
of the activation of reservoir units, as follows:
$\y(\mathbf{t}) = \Wout \x(t)$, where $\Wout \in \R^{N_Y \times N_L N_R}$ is the matrix of output weights. This 
combination
allows to differently weight the contributions of 
the multiple dynamics developed in the network's state. 
The training of the network is performed only on the readout layer by means of direct numerical methods.
Finally, as pre-training technique we use the Intrinsic Plasticity (IP) adaptation for deep recurrent architectures, particularly effective for DeepESN and ESN architectures \cite{gallicchio2017deep, DesignDeepESN}.

\section{Experimental Comparison}
In this section we present the results of the experimental comparison performed between randomized and fully-trained RNNs. The approaches are assessed on polyphonic music tasks defined in \cite{boulanger2012modeling}.
In particular, we consider the following 4 datasets\footnote{Piano-midi.de (\url{www.piano-midi.de}); MuseData (\url{www.musedata.org}); JSBchorales (chorales by J. S. Bach);   Nottingham (\url{ifdo.ca/~seymour/nottingham/nottingham.html}).
}: Piano-midi.de, MuseData~, JSBchorales and Nottingham. 
A polyphonic music task is 
defined as
a next-step prediction 
on 88-, 82-, 52- and 58- dimensional sequences for Piano-midi.de, MuseData, JSBchorales and Nottingham 
datasets,
respectively. 
Since these datasets consist in high-dimensional time-series characterized by heterogeneous sequences, sparse vector representations and complex temporal dependencies involved at different time-scales, they are considered challenging real-world benchmarks for RNNs \cite{bengio2013advances}.

Models' performance is measured by using 
the expected frame-level accuracy (ACC),
commonly adopted as prediction accuracy in polyphonic music tasks
\cite{boulanger2012modeling}, and computed as follows:
\begin{equation}
	\text{ACC} = \frac{\sum_{t=1}^{T} TP(t)}{\sum_{t=1}^{T} TP(t) + \sum_{t=1}^{T} FP(t) + \sum_{t=1}^{T} FN(t)},
\end{equation}
where $T$ 
is the total number of time-steps, while $TP(t)$, $FP(t)$ and $FN(t)$ 
respectively denote the numbers of 
true positive, false positive and false negative notes
predicted at time-step $t$.

Concerning DeepESN and ESN approaches, we considered reservoirs 
initialized with
$1\%$ of connectivity. 
Moreover, we performed a model selection on the major hyper-parameters considering spectral radius $\rho$ and leaky integrator $a$ values in $\{0.1, 0.3, 0.5, 0.7, 0.9, 1.0\}$, and input scaling $\sigma$ values in $\{0.5, 1.5, 2.5\}$.
 Training of the readout was performed through ridge regression \cite{Jaeger2004,lukovsevivcius2009reservoir} with regularization coefficient $\lambda_r$ in $\{10^{-4}, 10^{-3}, 10^{-2}, 10^{-1}\}$. Moreover, based on the results of the design analysis in \cite{DesignDeepESN} on polyphonic music tasks, we set up DeepESN with $N_L=30$ layers composed by $N_R=200$ units, and ESN with $N_R=6000$ recurrent units.
We used an IP adaptation configured as in \cite{gallicchio2017deep, DesignDeepESN} with a standard deviation of $\sigma_{IP} = 0.1$.

For what regards fully trained RNNs, we used the Adam learning algorithm \cite{kinga2015method} with a maximum of 2000 epochs. In order to regularize the learning process, we applied dropout methods, a clipping gradient with a value of $5$ and an early stopping with a patience value of $30$.  
Then, we performed a model selection considering learning rate values in $\{10^{-4}, 10^{-3}, 10^{-2}, 10^{-1}\}$ and dropout values in $\{0.1, 0.2,$ $0.3, 0.4, 0.5\}$.

Since randomized and fully-trained RNNs implement different learning approaches, it is difficult to set up a fair experimental comparison between them.
However, we faced these difficulties by considering a comparable number of free-parameters 
for all the models.
The number of recurrent units and free-parameters considered in the models is shown in the second and third columns of Tab.~\ref{tab.results}. 
Each model is individually selected on the validation sets through a grid search on hyper-parameters ranges. We independently generated 5 guesses for each network hyper-parametrization (for random initialization),
and averaged the results over such guesses.

\begin{table}[t]
\centering

\small
\setlength\tabcolsep{1.5pt}

\begin{tabular}{|l|l|l|l|l|}
\hline
Model & total recurrent units & free-parameters & test ACC & computation time \\
\hline
\multicolumn{5}{|c|}{Piano-midi.de}\\ 
\hline
DeepESN  & $6000$ & $540088$ & \textbf{33.33 (0.11)} \textbf{\%} & \textbf{386}  \\ 
ESN  & $6000$ & $540088$   & 30.43 (0.06) \textbf{\%} & 748 \\ 
SRN  & $652$ & $540596$   & 29.48 (0.35) \% & 3185  \\ 
LSTM  & $316$ & $539816$   & 28.98 (2.93) \% &  2333  \\ 
GRU  & $369$ & $539566$   & 31.38 (0.21) \% & 2821  \\ 
\hline
\multicolumn{5}{|c|}{MuseData}\\
\hline
DeepESN  & $6000$ & $504082$ & \textbf{36.32 (0.06)} \textbf{\%} & \textbf{789}  \\ 
ESN  & $6000$ & $504082$   & 35.95 (0.04) \textbf{\%} & 997  \\ 
SRN  & $632$ & $503786$  & 34.02 (0.28) \% & 8825  \\ 
LSTM  & $307$ & $504176$   & 34.71 (1.17) \% & 18274  \\ 
GRU  & $358$ & $503072$  & 35.89 (0.17) \% & 18104  \\  
\hline
\multicolumn{5}{|c|}{JSBchorales}\\
\hline
DeepESN  & $6000$ & $324052$   & \textbf{30.82  (0.12)} \textbf{\%} & \textbf{83}  \\ 
ESN  & $6000$ & $324052$  & 29.14  (0.09) \textbf{\%} & 140  \\ 
SRN  & $519$ & $323908$   & 29.68 (0.17) \% & 341  \\ 
LSTM  & $254$ & $325172$  & 29.80 (0.38) \% & 532  \\ 
GRU  & $295$ & $323372$   & 29.63 (0.64) \% & 230  \\ 
\hline
\multicolumn{5}{|c|}{Nottingham}\\
\hline
DeepESN  & $6000$ & $360058$ & 69.43 (0.05) \% & \textbf{677}  \\ 
ESN  & $6000$ & $360058$  & 69.12 (0.08) \% & 1473  \\ 
SRN  & $545$ & $360848$   & 65.89 (0.49) \% & 2252 \\ 
LSTM  & $266$ & $361286$  & 70.00 (0.24) \% & 26175  \\ 
GRU  & $309$ & $359116$   & \textbf{71.50 (0.77)} \% & 11844  \\  
\hline
\end{tabular}

\caption{Free-parameters and test ACC achieved by DeepESN, SRN, LSTM and GRU. Computation time represents the seconds to complete training and test.}
\label{tab.results}
\end{table}

In accordance with the different characteristics of the considered training approaches (direct methods for RC and iterative methods for fully-trained models)
we preferred the most efficient method
in all the considered cases.
Accordingly, we used a MATLAB implementation for DeepESN and ESN models, and a Keras implementation for fully-trained RNNs.
We measured the time in seconds spent by models in training and test procedures, performing experiments on a CPU 
``Intel Xeon E5, 1.80GHz, 16 cores''
in the case of RC approaches, and on a GPU 
``Tesla P100 PCIe 16GB''
in the case of fully-trained RNNs, 
with the same aim to give the best resource to each of them.

Tab.~\ref{tab.results} shows the number of recurrent units, the number of free-parameters, the predictive accuracy and the computation time (in seconds) achieved by DeepESN, ESN, SRN, LSTM and GRU models.
For what regards the comparison between RC approaches in terms of predictive performance, results indicate that
DeepESN outperformed ESN with an accuracy improvement of $2.90 \%$, $0.37 \%$, $1.68 \%$ and $0.31 \%$ on Piano-midi.de, MuseData, JSBchorales and Nottingha tasks, respectively. Concerning the comparison between fully-trained RNNs, GRU obtained a similar accuracy to SRN and LSTM models on JSBchorales task and it outperformed them on Piano-midi.de, MuseData and Nottingham tasks. 

The efficiency assessments show that DeepESN requires about less that one order of magnitude of computation time with respect to fully-trained RNNs,
boosting the already striking efficiency 
of standard ESN models.
Moreover, while ESN benefits in terms of efficiency only by exploiting the sparsity of reservoirs (with $1 \%$ of connectivity), in the case of DeepESN the benefit is intrinsically due to the architectural constraints involved by layering \cite{DesignDeepESN} (and are obtained also with fully-connected layers).
 
Overall, the DeepESN model outperformed all the other approaches on 3 out of 4 tasks, resulting extremely more efficient with respect to fully-trained RNNs.

\section{Conclusions}
In this paper, we performed an experimental comparison between radomized and fully-trained RNNs on challenging real-world tasks characterized by multivariate time-series. 
This kind of comparisons in complex temporal tasks, that is practically absent in literature especially for what regards efficiency aspects, offered the opportunity to assess efficient alternative models (ESN and DeepESN in particular) to typical RNN approaches (LSTM and GRU).
Moreover, we assessed also the effectiveness of layering in deep recurrent architectures with a large number of layers (i.e., 30).

Concerning fully-trained RNNs, GRU outperformed the other gated RNNs on 3 out of 4 tasks and it was more efficient than LSTM in most cases. The effectiveness of GRU approaches found in our experiments is in line with the literature that deals with the design of adaptive gates in recurrent architectures. 

For what regards randomized RNNs, the results show that DeepESN is able to outperform ESN in terms of prediction accuracy and efficiency on all tasks. Interestingly, this highlights that the layering aspect allows us to improve the effectiveness of RC approaches on multiple time-scales processing.
Overall, the DeepESN model outperformed other approaches in terms of prediction accuracy on 3 out of 4 tasks. Finally, DeepESN required much less time in computation time with respect to the others models resulting in an extremely efficient model able to compete with the state-of-the-art on challenging time-series tasks. 

More in general, it is interesting to highlight the gain in the prediction accuracy showed by the multiple time-scales processing capability obtained by layering in deep RC models and by using adaptive gates in fully-trained RNNs in comparison to the respective baselines (ESN and SRN, respectively).
Also, it is particularly interesting to note the comparison between models with the capability to learn multiple time-scales  dynamics (LSTM and GRU)
and models showing an intrinsic capability to 
develop such kind of hierarchical temporal representations
(DeepESN), which was completely lacking in literature.

In addition to provide insights on such general issues, this paper would 
contribute to show a practical way to efficiently approach the design of learning models in the scenario of deep RNN, extending the set of tools available to the users for complex time-series tasks. 
Indeed, the first empirical results provided in this paper seem to indicate that some classes of models are sometimes uncritically adopted, i.e. despite their cost, 
guided by the natural popularity due to their software availability (GRU, LSTM). 
The same diffusion of software tools deserve more effort on the side of the other models (DeepESN class), although the first instances are already available\footnote{DeepESN implementations are made publicly available for download both in $\textrm{MATLAB}$ (see \url{https://it.mathworks.com/matlabcentral/fileexchange/69402-deepesn}) and in Python (see \url{https://github.com/lucapedrelli/DeepESN}).}.

\begin{footnotesize}
\bibliographystyle{unsrt}
\bibliography{references}

\end{footnotesize}

\end{document}